\documentclass[11pt,a4paper]{article}
\usepackage[hyperref]{wmt2020}
\usepackage{times}
\usepackage{latexsym}

\usepackage{microtype}
\usepackage{amsfonts,amssymb}
\usepackage{graphicx}
\usepackage{amsmath}
\usepackage{multirow}
\usepackage{booktabs}
\usepackage{amsfonts}
\usepackage{algorithm}
\usepackage{algorithmicx, algpseudocode}
\usepackage{mathtools}
\usepackage{arydshln}
\usepackage{multicol}

\aclfinalcopy 

\title{{SJTU-NICT}{'}s Supervised and Unsupervised Neural Machine Translation Systems for the {WMT}20 News Translation Task}

\author{
	Zuchao Li$^{1,2,3}$,
	Hai Zhao$^{1,2,3,}$\thanks{$\ $ Corresponding authors. This paper was partially supported by National Key Research and Development Program of China (No. 2017YFB0304100), Key Projects of National Natural Science Foundation of China (U1836222 and 61733011), Huawei-SJTU Long Term AI Project, Cutting-edge Machine Reading Comprehension and Language Model. Rui Wang was partially supported by JSPS grant-in-aid for early-career scientists (19K20354): ``Unsupervised Neural Machine Translation in Universal Scenarios" and NICT tenure-track researcher startup fund ``Toward Intelligent Machine Translation".},\\
	\textbf{Rui Wang$^{4,*}$,}
	\textbf{Kehai Chen$^{4}$,}
	\textbf{Masao Utiyama$^{4}$,}
	and \textbf{Eiichiro Sumita}$^{4}$
	\\
	$^1$Department of Computer Science and Engineering, Shanghai Jiao Tong University (SJTU)\\
	$^2$Key Laboratory of Shanghai Education Commission for Intelligent Interaction\\
	and Cognitive Engineering, Shanghai Jiao Tong University, Shanghai, China\\
	$^3$MoE Key Lab of Artificial Intelligence, AI Institude, Shanghai Jiao Tong University, China\\
	$^4$National Institute of Information and Communications Technology (NICT), Kyoto, Japan \\
	 {\tt charlee@sjtu.edu.cn, zhaohai@cs,sjtu.edu.cn, wangrui@nict.go.jp} \\
}

\date{}

\begin{document}
\maketitle
\begin{abstract}

In this paper, we introduced our joint team SJTU-NICT 's participation in the WMT 2020 machine translation shared task. In this shared task, we participated in four translation directions of three language pairs: English-Chinese, English-Polish on supervised machine translation track, German-Upper Sorbian on low-resource and unsupervised machine translation tracks. Based on different conditions of language pairs, we have experimented with diverse neural machine translation (NMT) techniques: document-enhanced NMT, XLM pre-trained language model enhanced NMT, bidirectional translation as a pre-training, reference language based UNMT,  data-dependent gaussian prior objective, and BT-BLEU collaborative filtering self-training. We also used the TF-IDF algorithm to filter the training set to obtain a domain more similar set with the test set for finetuning. In our submissions, the primary systems won the first place on English to Chinese, Polish to English, and German to Upper Sorbian translation directions. 

\end{abstract}

\section{Introduction}

Our SJTU-NICT team participated in the WMT20 shared task, including supervised track, unsupervised, and low-resource track. During the participation, we placed our attention on Polish (PL) $\rightarrow$ English (EN) and English (EN) $\rightarrow$ Chinese (ZH) on the supervised track, while on the unsupervised and low-resource track, the German (DE) $\leftrightarrow$ Upper Sorbian (HSB) both directions are focused.

Our  baseline system in supervised track is based on the Transformer big architecture proposed by \citet{vaswani2017attention}, in which its open-source implementation version {\tt Fairseq} \cite{ott2019fairseq} is adopted. In the unsupervised and low-resource track, we draw on the successful experience of the {\tt XLM} framework \cite{conneau2019unsupervised}, and used the two-stage training mode of masked language modeling (MLM) pre-training + back-translation (BT) finetune to obtain a very strong baseline performance. {\tt Marian} \cite{junczys2018marian} toolkit is utilized for training the decoder in reranking using machine translation targets instead of common GPT-style language modeling targets.

In order to better play the role of WMT evaluation in polishing the methods proposed or improved by our team \cite{he-etal-2018-syntax,li-etal-2018-seq2seq,zhang-etal-2018-modeling,zhang2018minimum,xiao-etal-2019-lattice,zhou-zhao-2019-head,li2019cross,luo-zhao-2020-bipartite}, we divided the three language pairs we participated in into three categories: 

\noindent 1. Traditional language pair with rich parallel corpus: EN-PL,

\noindent 2.  Language pair with document-level information: EN-ZH,

\noindent 3.  Language pair with no or low parallel resources: DE-HSB.

In the supervised PL$\rightarrow$EN translation direction, we based on the XLM framework to pre-train a Polish language model using common crawl and news crawl monolingual data, and proposed the XLM enhanced NMT model inspired from the idea of incorporating BERT into NMT \cite{zhu2020incorporating}. Besides, we trained a bidirectional translation model of EN-PL based on the parallel corpus and further finetuned it to the PL$\rightarrow$EN direction.

In the supervised EN$\rightarrow$ZH translation with document information, we propose a document enhanced NMT model based on Longformer \cite{Beltagy2020Longformer}. The training of our proposed document enhanced NMT model is split into three stages.  In the first stage, we pre-train the Longformer document encoder with MLM target on the document text in Wikipedia dumps, UN News, and News Commentary monolingual corpus. A conventional Transformer-big NMT model is trained in the second stage. In the final stage, the Longformer encoder and conventional Transformer big NMT model are used to initialize the full document-enhanced NMT model parameters, in which the Longformer encoder is adopted to extract representations for the document of an input sequence, and then the document representations are fused with each layer of the encoder and decoder of the NMT model through attention mechanisms. 

In the unsupervised machine translation track on DE-HSB, we experimented with the reference language based UNMT (RUNMT) \cite{li2020reference} framework we proposed recently. Under this framework, we choose English as the reference language, and use the Europarl parallel corpus of EN-DE to enhance the unsupervised machine translation between DE and HSB. Specifically, we adopted reference language translation (RAT), reference language back-translation (RABT), and cross-lingual back-translation (XBT) three training targets with the help of the cross-lingual agreement provided by the EN-DE parallel corpus to enhance the unsupervised translation performance.

Due to the introduction of more explicit supervision signals brought by parallel corpus in the low-resource machine translation track on DE-HSB, we discarded the use of the weaker agreement provided by the reference language,  conducted joint training on the unsupervised back-translation and the supervised (forward-)translation directly, and introduced BT-BLEU based collaborative filtering technology for further self-training. In addition, inspired by our previous work \cite{sun2020unsupervised}, we also use MLM and translation language modeling (TLM) to continue pre-training the model while machine translation training.

In addition, in all basic NMT models, we empower the training process with our proposed data-dependent gaussian prior objective (D2GPo) \cite{Li2020Data-dependent}, so that the model can maintain the diversity of the output. When the main model training is finished, the TF-IDF algorithm is employed to filter the training set according to the input of the test set, a training subset whose domain is more similar to the test set is obtained, and then used to finetune the model for reducing the performance degradation caused by domain inconsistency. For the final submission, an ensemble of several different trained models outputs the $n$-best predictions, and used the decoder trained with {\tt Marian} toolkit to performs reranking to get the final system output.

\section{Methodology}

\subsection{XLM-enhanced NMT}\label{subsec:xlm-nmt}

Pre-trained language models such as ELMo \cite{peters-etal-2018-deep}, BERT \cite{devlin-etal-2019-bert}, XLM \cite{conneau2019unsupervised}, XLNet \cite{yang2019xlnet}, ALBERT \cite{lan2019albert} etc. have recently demonstrated a very dominant effect on natural language processing tasks. Several works \cite{clinchant2019use,imamura2019recycling,zhu2020incorporating} leveraged a pre-trained BERT model for improving NMT and found that BERT can bring significantly better results over the baseline. 

Since BERT and other pre-trained language models are trained on large scale corpus beyond the data provided by the WMT20 organizers, the direct use of BERT will make the system submitted unconstrained. Using an XLM  model, a variant of BERT, pre-trained from scratch on the monolingual data provided by the official to enhance our NMT model, is a good choice to keep the system constrained. Moreover, the XLM model has the advantages of simple training preprocessing, low requirement for training environment that no specialized hardware such as TPU is needed. Inspired by the \textit{BERT-fused model} proposed by \citet{zhu2020incorporating}, we built a \textit{XLM-enhanced model}, in which we utilize XLM context-aware representations to adaptively interact with all layers in the NMT model with attention mechanism, instead of serving it as input embeddings only.

In the \textit{XLM-enhanced model}, XLM as an additional encoder and the original encoder of NMT constitute a dual-encoder structure, which is very similar to our previous work \cite{li2019explicit}. The XLM-encoder attention and XLM-decoder attention are essentially the same with the Representation Learning Frameworks (RLFs) we proposed: Source-side fusion RLF (SRLF), Target-side fusion RLF (TRLF), and both-side fusion RLF (BRLF, which is a combination of SRLF and TRLF). Specifically, in the SRLF, given a source language input $x$, a Pre-trained Language Modeling (PLM) encoder (like BERT, XLM) first encodes it into a context-aware representation:
\begin{equation}
H_P = {\tt PLM}^k(x),
\end{equation}
where $H_P$ is the output of the $k$-th layer of the PLM encoder. As PLM and NMT models adopt different sub-word segmentation rules or algorithms and the addition of special tokens are different, the input sequence length of PLM and NMT encoders is inconsistent or cannot correspond in every position. Assuming that $i$ represents the position of the input sequence of NMT encoder, the hidden state $H_E^l$ after fusion with $H_P$ in SRLF of the $l$-th layer is:
\begin{equation}
\begin{split}
H_E^l = \frac{1}{2}({\tt attn_S}(H_E^{l-1}, H_E^{l-1}, H_E^{l-1}) \\
+ {\tt attn_P}(H_E^{l-1}, H_P, H_P)),
\end{split}
\end{equation}
where ${\tt attn}_S$ is a multi-head self-attention layer and ${\tt attn}_P$ is the multi-head attention layer. $H_E$ will eventually be output from the last layer as the final representation.

In the TRLF framework,  the dual-encoder provides two encoded outputs; the decoder will use both contexts at the same time. In the case of layer $l$ in the decoder, we have
\begin{equation}
\begin{split}
H_{DS}^l = {\tt attn_{MS}}(H_D^{l-1}, H_D^{l-1}, H_D^{l-1}), \\
H_D^l = \frac{1}{2}({\tt attn_{EC}}(H_{DS}^l, H_E, H_E) \\
+{\tt attn_{PC}}(H_{DS}^l, H_P, H_P)),
\end{split}
\end{equation}
where ${\tt attn_{MS}}$ is the multi-head future-masked self-attention layer, ${\tt attn_{EC}}$ and ${\tt attn_{PC}}$ are independent multi-head attention layer for context query.

In the condition that SRLF framework is only used, the representation of PLM is only fused into the final representation $H_E$ in the encoder side; then the decoder side continues to use the original decoding ways: $H_D^l = {\tt attn_{PC}(H_{DS}^l, H_E, H_E)}$. While the the TRLF framework is only adopted,  the output of NMT encoder is $H_E = {\tt attn_S(H_E^{l-1}, H_E^{l-1}, H_E^{l-1})}$. A BRLF framework is a combination of these two frameworks.

Moreover, in the training of the RLFs, a same drop-net trick proposed by \citet{zhu2020incorporating} is adopted to ensure that the features output by PLM and the conventional encoder are fully utilized. In this method, the interval of 0-1 is divided into three parts according to the pre-set drop-net ratio $p_{net}$, where $[0, \frac{p_{net}}{2})$ is the probability of attending to the final sum for the first ${\tt attn}$ in $H_E^L$ and $H_D^L$, $[\frac{p_{net}}{2}, 1-\frac{p_{net}}{2})$ is the probability for the whole $H_E^L$ and $H_D^L$ equation, $[1-\frac{p_{net}}{2}, 1]$ is the probability for the second ${\tt attn}$ in in $H_E^L$ and $H_D^L$.

\subsection{Bidirectional NMT}

Machine translation, in general, is unidirectional, that is, from the source language to the target language. The encoder-decoder framework for NMT has been shown effective in large data scenarios, and the more high-quality bilingual training data, the better performance the model tends to achieve. Recent works \cite{zoph2016transfer, kim2019effective} on translation transfer learning \cite{torrey2010transfer, pan2009survey} from rich-resource language pairs to low-resource language pairs demonstrate that translation has some universal nature in essence between different language pairs. As the source-to-target (S2T) forward translation and target-to-source (T2S) backward translation can be seen as two special language pairs in bilingual translation, it can make use of the translation universal nature to improve each other, i.e., dual learning \cite{he2016dual}. Based on this motivation, we developed a bidirectional NMT model, in which the S2T and T2S translation were trained and optimized jointly. Therefore, the training data was doubled to make better and full use of the costly bilingual corpus.

Given parallel corpus $\mathcal{C} = \{(x^{(n)}, y^{(n)})\}_{n=1}^N$, the bidirectional NMT model is trained in two phase. In the first \textit{bidirectional translation as pre-training} phase, a joint training objective is used to jointly maximize the likelihood of both translation direction on the bilingual data:
\begin{equation}
\mathcal{L}(\theta_{parent})=\sum_{n=1}^N (\log p(y^{(n)}|x^{(n)}) + \log p(x^{(n)}|y^{(n)})),
\end{equation}
where $\theta_{parent}$ is the parameters of the model, namely \textit{parent model}, obtained in this phase.

The second phase is \textit{unidirectional translation fine-tuning}. Although there are commonalities in different translation directions, the differences are also very obvious. To further expose the model to the direction difference and improve the effect of unidirectional translation, we further finetune the bidirectional pre-trained model on the bilingual data. Take S2T translation as an example; the model is optimized as follows:
\begin{equation}
\mathcal{L}(\theta_{S \rightarrow T}) = \sum_{n=1}^N \log p(y^{(n)}|x^{(n)}),
\end{equation}
where $\theta_{S \rightarrow T}$ is the parameters of \textit{child model} which is initialized with $\theta_{parent}$. Similarly, the T2S \textit{child model} can also be obtained.

Due to the introduction of bidirectional translation in one model, follow the practice of \citet{conneau2019cross}, shared sub-word vocabulary and shared encoder-decoder (source and target) embedding were employed to improves the alignment of embedding spaces across languages. In addition, since the encoder and decoder need to be able to handle two languages simultaneously, a language embedding was used to indicate the language being processed, so as to reduce confusion of the model.

\subsection{Document-enhanced NMT}

In spite of its success \cite{vaswani2017attention}, sentence-level NMT has been based on strong independence and locality assumptions generally, in which the interrelations among these discourse \cite{jurafsky2000speech} elements were ignored. This results in that the translations may be perfect at the sentence-level but lack crucial properties of the text, hindering understanding \cite{maruf2019survey}. To help to resolve ambiguities and inconsistencies in translations, some MT pioneers \cite{bar1960present, xiong2013lexical, sennrich2018time} exploit the underlying discourse structure information of a text to address this issue, while others \cite{bawden2018evaluating, voita2018context, jean2019context, wang2019improving, scherrer2019analysing} extend the translation units with the context or use an additional context encoder and attention. It is worth noting that the essence of the document-level NMT claimed with additional context and attention is still sentence-level MT, whose translation is still output sentence by sentence. We named it as document-enhanced NMT more precisely.

Due to computational efficiency and tractability concerns, the document-enhanced NMT models mostly used document embedding, document topic information, and limited past or future context sentences, etc., rather than the truly whole document information. Recently, with the increase in computational power available to us and the well-designed neural network structures \cite{dai2019transformer,kitaev2019reformer,Beltagy2020Longformer} for long sequence encoding, we are finally in a position to employ the whole document information for enhancing sentence-level NMT. In addition, we argue that since long sequences encoding is easier than decoding, truly whole document-level translation is still a long way off, since the bidirectional context is available in the encoder, but only the past is visible by the decoder.

\paragraph{Longformer} To make the long documents processed with Transformer \cite{vaswani2017attention} architecture feasible or easier, a modified Transformer architecture named Longformer was proposed by \citet{Beltagy2020Longformer}, in which the limitation for memory and computational requirements is addressed with a novel self-attention operation scales linearly with the sequence length.

In Longformer, the original full self-attention ($O(n^2)$ time and memory complexity) is sparsified to makes it efficient for longer sequences. There are three ``attention patterns" for specifying pairs of input locations attending to one another.
\begin{itemize}
	\item \textbf{Sliding Window} Self-attention is performed in a fixed-size window $w$ and multiple stacked layers of such sliding windowed attention results in a large receptive field as analogs to CNNs.
	\item \textbf{Dilated Sliding Window} Inspired by the dilated CNNs \cite{oord2016wavenet}, dilation gaps of size $d$ is introduced to the window to further increase the receptive field without increasing computation.
	\item \textbf{Global Attention} Though the receptive field is enlarged by stacking multiple layers and dilation in sliding window and dilated sliding window attention patterns, some part of the long sequence has the requirement for keeping the full and global receptive field due to the downstream tasks, so global attention is introduced to make up this need.
\end{itemize}

In our document-enhanced NMT model, some heads in multi-head attention are set to use the sliding window pattern to focus on the local context which was revealed very important \cite{kovaleva2019revealing}, while others with dilation focus on longer context. Besides, as Longformer is incorporated into the NMT model, we perform global attention on the position of {\tt [CLS]} token in which the representation of the whole sequence (i.e., the document embedding) is generated. This makes the previous document-enhanced model with document embedding as a special case of ours. It is worth noting that since the sentence being translated is part of the document, setting its positions in the document to use global attention pattern will improve the performance; but to reduce the document computation and use cache for acceleration (not recalculate the document for each sentence), we only attend the {\tt [CLS]} position globally.

In our document-enhanced model, the Longformer is first pre-trained with the masked language modeling objective on the monolingual document corpus. It is fixed throughout the NMT training to reduce the model parameters optimized in the training stage. Thus, Longformer can also be thought of as a pre-trained language model, as it provides a document context representation $H_P$ for the NMT model, the integration of Longformer in \textit{Document-enhanced NMT} is consistent with the XLM model in \textit{XLM-enhanced NMT}.

\subsection{Reference Language based UNMT}

The rise of UNMT almost completely relieves the parallel corpus curse, though UNMT is still subject to unsatisfactory performance due to the vagueness of the clues available for its core back-translation training.  Further enriching the idea of pivot translation by extending the use of parallel corpora beyond the source-target paradigm, we propose a new reference language-based framework for UNMT, RUNMT, in which the reference language only shares a parallel corpus with the source, but this corpus still indicates a signal clear enough to help the reconstruction training of UNMT through a proposed reference agreement mechanism.

Specifically, we proposed three kinds of reference agreement utilization approaches in \cite{li2020reference}: reference agreement translation (RAT), reference agreement back-translation (RABT), and cross-lingual back-translation (XBT).

\paragraph{RAT} RAT utilizes the principle for translating paired sentences into the target language $\mathcal{T}$ of the source $\mathcal{S}$ and reference $\mathcal{R}$ language. Since the input the parallel, the both translation outputs should be the same. Given a parallel sentence pair $\langle s, r \rangle$ between language $\mathcal{S}$ and $\mathcal{R}$, we would ideally have $\mathbb{P}(\cdot|s; \theta_{\mathcal{S} \rightarrow \mathcal{T}}) = \mathbb{P}(\cdot|r; \theta_{\mathcal{R} \rightarrow \mathcal{T}})$, where $\theta_{\mathcal{S} \rightarrow \mathcal{T}}$ and $\theta_{\mathcal{R} \rightarrow \mathcal{T}}$ represent $\mathcal{S} \rightarrow \mathcal{T}$ and $\mathcal{R} \rightarrow \mathcal{T}$ translation models respectively. However, as the two models  are trained on different data, the agreement may be corrupted. Therefore, we combine the two models  to obtain the agreed-upon translation output $\tilde{t}_a$:
\begin{equation}
\tilde{t}_a \sim \mathbb{P}(\cdot|s, r; \theta_{\mathcal{S} \rightarrow \mathcal{T}}, \theta_{\mathcal{R} \rightarrow \mathcal{T}}),
\end{equation}
where $\mathbb{P}(\cdot|s, r; \theta_{\mathcal{S} \rightarrow \mathcal{T}}, \theta_{\mathcal{R} \rightarrow \mathcal{T}})$ is
\begin{equation}
\prod_{i=1}^{J} [ \frac{1}{2}(\mathbb{P}(\cdot|s, \tilde{t}_{<i}; \theta_{\mathcal{S} \rightarrow \mathcal{T}}) + \mathbb{P}(\cdot|r, \tilde{t}_{<i}; \theta_{\mathcal{R} \rightarrow \mathcal{T}}))],
\end{equation}
$\tilde{t}_{<i}$ indicates the decoded tokens before the $i$-the generation step.

Finally, two synthetic sentence pairs $\langle s,\tilde{t}_a \rangle$ and $\langle r,\tilde{t}_a \rangle$ are used to train the models $\mathcal{S} \rightarrow \mathcal{T}$ and $\mathcal{R} \rightarrow \mathcal{T}$. Since the silver learning target is optimized, the smoothed cross-entropy loss $\mathcal{L}_\epsilon$ is used instead of the ordinary cross-entropy loss $\mathcal{L}$. The learning objective for RAT can be written as:
\begin{equation}
\footnotesize
\mathcal{L}_{\textbf{RAT}}(\mathcal{S}, \mathcal{T}, \mathcal{R}) = \mathcal{L}_\epsilon(\theta_{\mathcal{S} \rightarrow \mathcal{T}}) + \mathcal{L}_\epsilon(\theta_{\mathcal{R} \rightarrow \mathcal{T}}),
\end{equation}

\paragraph{RABT} With the regularized pseudo-parallel sentences in RAT, we not only train the $\mathcal{S} \rightarrow \mathcal{T}$ and $\mathcal{R} \rightarrow \mathcal{T}$ forward-translation models (as the generation direction is the same as the training direction), but also train the BT models, i.e., $\mathcal{T} \rightarrow \mathcal{S}$ and $\mathcal{T} \rightarrow \mathcal{R}$. The learning objective of RABT can be described as:
\begin{equation}
\mathcal{L}_{\textbf{RABT}}(\mathcal{S}, \mathcal{T}, \mathcal{R}) = \mathcal{L}(\theta_{\mathcal{T} \rightarrow \mathcal{S}}) + \mathcal{L}(\theta_{\mathcal{T} \rightarrow \mathcal{R}}).
\end{equation}

\paragraph{XBT} The parallel corpus between languages $\mathcal{S}$ and $\mathcal{R}$ can not only bring agreement in the translations of the same target language $\mathcal{T}$, but also cross-lingual agreement, that is, using the target language as the bridge to form pivot translation \cite{wu2007pivot,utiyama2007comparison,paul2009importance} patterns: $\mathcal{S} \rightarrow \mathcal{T} \rightarrow \mathcal{R}$ and $\mathcal{R} \rightarrow \mathcal{T} \rightarrow \mathcal{S}$.  In XBT, paired sentences $s$ and $r$ are translated to language $\mathcal{T}$: $\tilde{t}_s$ and $\tilde{t}_r$, and forms two new pseudo-parallel pairs: $\langle \tilde{t}_s, r \rangle$ and $\langle \tilde{t}_r, s \rangle$, which promote the training of translation $ \mathcal{T} \rightarrow \mathcal{R}$ and $\mathcal{T} \rightarrow \mathcal{S}$. The objective function of XBT is:
\begin{equation}
\mathcal{L}_{\textbf{XBT}}(\mathcal{S}, \mathcal{T}, \mathcal{R}) = \mathcal{L}(\theta_{\mathcal{T} \rightarrow \mathcal{R}}) +  \mathcal{L}(\theta_{\mathcal{T} \rightarrow \mathcal{S}}),
\end{equation}

\subsection{CFST: Collaborative Filter for Self-Training with BT-BLEU}

Self-training, proposed by \citet{scudder1965probability}, is a
semi-supervised approach that utilizes unannotated data to create better models. Recently, self-training has been successfully applied to both NMT and UNMT fields \cite{he2019revisiting,sun2020self}, especially for the unbalanced low-resource training data scenarios.

Formally, in self-training strategy for machine translation, a parallel dataset  $\mathcal{C} = \{(x^{(n)}, y^{(n)})\}_{n=1}^N$ in NMT and a unpaired monolingual dataset $\mathcal{D} = \{x^{(m)}\}_{m=1}^M \cup \{y^{(n)}\}_{n=1}^N$ in UNMT is used to train the initial model. Then, a subset of pseudo parallel data is incorporated to update the model with a pseudo-supervised NMT (PNMT) objective (including forward translation and backward translation) for both NMT and UNMT as shown in Algorithm \ref{alg:st}.  In NMT, a large unlabeled dataset $\mathcal{U} =\{x^{(j)}\}_{j=1}^{L}$ is used for the synthesis of pseudo-parallel corpora. While in UNMT, since the model is trained with back-translation on unpaired monolingual data, the pseudo-parallel corpora is synthesized by the monolingual data, i.e., $\mathcal{U} =\{x^{(m)}\}_{m=1}^{M}$.

\begin{algorithm}[!t]
	\centering
	\caption{Classic Self-training}
	\label{alg:st}
	\begin{algorithmic}[1]
		\State Train a base NMT/UNMT model $f_{\theta_{S \rightarrow T}}$ on $\mathcal{C}$ 
		\Repeat
		\State Apply $f_{\theta_{S \rightarrow T}}$ to the unlabeled instances $\mathcal{U}$
		\State Select a subset $\mathcal{Q} \subset \{(x, f_{\theta_{S \rightarrow T}}(x)) | x \in \mathcal{U}\}$
		\State Update model $f_{\theta_{S \ \rightarrow T}}$ on $\mathcal{Q}$ with self-training objective and $\mathcal{C}$ with original objective
		\Until{convergence or maximum iterations are reached}
	\end{algorithmic}
\end{algorithm}

Considering the translation quality can't effectively be evaluated across languages in machine translation with only the monolingual data, therefore the selection of the subset $\mathcal{Q}$, is one of the key factors for self-training. It is usually selected based on some confidence scores (e.g. log probability or perplexity, PPL) \cite{yarowsky1995unsupervised}, but it is also possible for S to be the whole pseudo parallel data \cite{zhu2009introduction}. In the backward translation based on the pseudo-parallel data, the DAE method widely used in UNMT can alleviate the impact of the noise resulted from the synthesized sentences on model training, since the synthesized sentences are only used as input. However, in the forward translation training, the quality of noisy targets will directly affect the success of the model training. Therefore, the selection of synthetic parallel corpus becomes particularly critical.

\begin{algorithm}[!t]
	\centering
	\caption{BT-BLEU based Collaborative Filter}
	\label{alg:btbleu}
	\begin{algorithmic}[1]
		\State Split $\mathcal{U}$ equally into two subsets $\mathcal{U}_1 = \{x^{(j)}\}_{j=1}^{L/2}$ and $\mathcal{U}_2 = \{x^{(j)}\}_{j=L/2+1}^L$ 
		\State Apply $f_{\theta_{S \rightarrow T}}$ to the unlabeled instances $\mathcal{U}_1$ and $\mathcal{U}_2$
		\State Train two backward translation models $f_{\theta_{T \rightarrow S}}^{(1)}$ with $\{(f_{\theta_{S \rightarrow T}}(x), x) | x \in \mathcal{U}_1\}$ and $f_{\theta_{T \rightarrow S}}^{(2)}$ with $\{(f_{\theta_{S \rightarrow T}}(x), x) | x \in \mathcal{U}_2\}$ respectively
		\State Translate $\{f_{\theta_{S \rightarrow T}}(x) | x \in \mathcal{U}_2\}$ with model $f_{\theta_{T \rightarrow S}}^{(1)}$, while $\{f_{\theta_{S \rightarrow T}}(x) | x \in \mathcal{U}_1\}$ with model $f_{\theta_{T \rightarrow S}}^{(2)}$
		\State Calculate BT-BLEU $\mathcal{B}$ for two subsets: $\textbf{BLEU}(f_{\theta_{T \rightarrow S}}^{(2)}(f_{\theta_{S \rightarrow T}}(x)), x), \forall x \in \mathcal{U}_1$ and $\textbf{BLEU}(f_{\theta_{T \rightarrow S}}^{(1)}(f_{\theta_{S \rightarrow T}}(x)), x), \forall x \in \mathcal{U}_2$
		\State $\mathcal{Q} = \{(x, f_{\theta_{S \rightarrow T}}(x)) | x \in \mathcal{U}_1, \mathcal{B} > \gamma\} \cup \{(x, f_{\theta_{S \rightarrow T}}(x)) | x \in \mathcal{U}_2, \mathcal{B} > \gamma\}$
	\end{algorithmic}
\end{algorithm} 

We propose a collaborative filtering algorithm based on BT-BLEU to select high quality pseudo-parallel pairs, as shown in Algorithm \ref{alg:btbleu}.  The BT-BLEU, as defined in \cite{li2020reference},  is a BLEU of $\textbf{\textit{x}} \in \mathcal{S}$ and $\tilde{\textbf{\textit{x}}}$ generated in the $\mathcal{S} \rightarrow \mathcal{T} \rightarrow \mathcal{S}$ back-translation process. As long as the model of $\mathcal{T} \rightarrow \mathcal{S}$ is fixed and the preference for translation of certain sentences is reduced as much as possible, BT-BLEU can reflect the translation quality of $\mathcal{S} \rightarrow \mathcal{T}$ to some extent, because of the necessary but insufficient condition that only the better the translation of $\mathcal{S} \rightarrow \mathcal{T}$ is, the better the translation of $\mathcal{T} \rightarrow \mathcal{S}$ can be.

To achieve the goal of reducing translation preferences, we split the pseudo parallel set into two subsets, ensure no overlap between two subsets. The model trained on subset 1 is used for back-translation on the subset 2, while the model on subset 2 back-translate the subset 1. This collaborative translation process enables the two models not to see the sentences to be translated, which guarantees the translation not relies on tricks. Additionally, we found that the sentences in different lengths have different difficulties for back-translation; we further divide the sentences into different bags according to their lengths and use different BT-BLEU threshold $\gamma$ for filtering.

\begin{table}[t]
	\centering
	\scalebox{1.0}{
		\begin{tabular}{lccc}
			\toprule
			
			\multirow{2}{*}{Systems} & Dev & \multicolumn{2}{c}{Test} \\
			\cmidrule{3-4}& BLEU & BLEU & chrF \\
			\midrule
			\multicolumn{3}{l}{\textbf{\textit{Base Data:}}} \\
			Transformer big & 25.8 & - & - \\
			XLM-enhanced & 26.8 &  - & -\\
			\midrule
			\multicolumn{3}{l}{\textbf{\textit{Base Data + ParaCrawl:}}} \\
			Transformer big & 30.0  & 32.2 & 0.596 \\
			\hdashline
			+D2GPo & 30.9 & - & - \\
			XLM-enhanced & 31.4 & - & -\\
			Bidirectional NMT & 29.5 & - & - \\
			\quad+Finetune & 31.2 & - & - \\
			\hdashline
			Ensemble & 32.0 & 34.0 & 0.606\\
			\quad++TF-IDF finetune & 32.3 & 34.2 & 0.609\\
			\quad++Re-ranking & 32.5 & 34.6 & 0.610\\
			\bottomrule
	\end{tabular}}
	\caption{PL$\rightarrow$EN performance (sacreBLEU and chrF score) for different models.}
	\label{tab:plen_bleu}
\end{table}

\subsection{TF-IDF Finetune}

NMT has been prominent in many machine translation tasks. However, in some domain-specific tasks, only the corpora from similar domains can improve translation performance. If a trained NMT model is evaluated on a domain mismatch corpus, the translation performance may even degrade. Therefore, domain adaptation techniques are essential to solve the NMT domain problem. It is a very common domain adaptation approach to further finetune the translation model trained on the domain-mixed corpus by using data that is the same or similar to the test set in domain. Therefore, we need to select sentences that are as close to the input domain as possible in the domain-mixed training set.

We argue that low-frequency words contain more domain information than high-frequency words, since low-frequency words are mostly domain-specific nouns, etc., which may indicate the topic directly. Therefore, we adopt the TF-IDF algorithm to search and filter on the whole training set. In fact, the improved version of TF-IDF algorithm, BM25 \cite{robertson2009probabilistic}, is employed to calculate the sentence similarity. BM25 is based on probabilistic information retrieval theory, whose score for a term $q$ to a sequence $Q$ is:
\begin{equation}
s(Q, q) = \frac{\textnormal{IDF} * ((k+1) * \textnormal{TF})}{(k * (1.0 - b + b * \frac{L_Q}{L_{\textnormal{avg}}}) + \textnormal{TF})},
\end{equation}
where $\textnormal{IDF}$ is the Inverse Document Frequency for term $q$ appears in the whole corpus, $\textnormal{TF}$ is the Term Frequency for $q$ in $D$, $L_Q$ represents the sequence length, $L_{\textnormal{avg}}$ is the average length of corpus $D$, $k$ and $b$ is the adjustable parameters.

With this scorer, every sequence will obtain a BM25 vector on the terms of the corpus:
\begin{equation}
V = [s(Q, t), \quad \forall t \in D_{\textnormal{terms}}],
\end{equation}
where $D_{\textnormal{terms}}$ indicates the all terms set in corpus $D$.  We calculate the cosine similarity as final scores between the query and every source sentence in corpus, and ranked on the scores to get the top-K pairs (K=1000 in our experiments) as the sub-training set for finetuning.

\section{Data Preprocessing and Model Setup}

Before model training, we preprocessed the data uniformly and customized the processing according to the requirements of each model. We normalized punctuation, remove non-printing characters, and tokenize all data with the Moses tokenizer \cite{koehn2007moses} except for the Chinese. For Chinese, we removed the segmentation space in some training data and then use PKUSeg \cite{pkuseg} toolkit to cut all Chinese sentences, so as to obtain unified word segmentation annotations. We use joint byte pair encodings (BPE) with 40K split operations for subword segmentation \cite{sennrich2016neural}.

In \textit{XLM-enhanced NMT} and \textit{Document-enhanced NMT}, we first train a basic NMT (Transformer big) model on the sentence-level data until convergence, then initialize the encoder and decoder of the \textit{XLM-enhanced NMT} and \textit{Document-enhanced NMT} full model with the obtained model. The PLM-encoder attention ${\tt attn_{P}}$ and PLM-decoder attention ${\tt attn_{PC}}$ are randomly initialized.

\paragraph{EN-PL}
On the language pair EN-PL, we explored performance in two training data settings. The first is \textit{base data}, including Europarl v10, Tilde Rapid corpus, and WikiMatrix bitext data, whose raw data is on the sentence-level. In the second setting \textit{base data + paracrawl}, we converted the paragraph-level alignment data in Paracrawl to sentence-level alignment and incorporated it with the \textit{base data}. In the conversion process, we adopted the method and program proposed by \cite{gale1993program} for aligning sentences based on a simple statistical model of character lengths, which uses the fact that longer sentences in one language tend to be translated into longer sentences in the other language, and that shorter sentences tend to be translated into shorter sentences. A probabilistic score is assigned to each proposed correspondence of sentences, based on the scaled difference of lengths of the two sentences (in characters) and the variance of this difference. This probabilistic score is used in a dynamic programming framework to find the maximum likelihood alignment of sentences.

For the Polish pre-trained XLM language model, we used all NewsCrawl monolingual data and some CommonCrawl monolingual data. Since the CommonCrawl data is very large and noisy and can potentially decrease the performance of LM if it is used in its raw form. We apply language identification filtering ({\tt langid}; \citet{lui2012langid}), keeping sentences with correct languages. In order to filter out the sentences shorter than 5 words or longer than 150 words more precisely, we re-split sentences using Spacy \cite{honnibal2017spacy} toolkit.

\begin{table}[t]
	\centering
	\scalebox{1.0}{
		\begin{tabular}{lccc}
			\toprule
			
			\multirow{2}{*}{Systems} & 19test & \multicolumn{2}{c}{Test} \\
			\cmidrule{3-4}& BLEU & BLEU & chrF \\
			\midrule
			Transformer big & 37.2 &  - & -  \\
			\hdashline
			+D2GPo & 37.7 &  - & - \\
			XLM-enhanced & 38.9 &  - & - \\
			Document-enhanced & 39.2 & - & - \\
			\hdashline
			Ensemble & 40.0 & 48.6 & 0.418 \\
			\quad++TF-IDF finetune & 40.2 &  48.8 & 0.422 \\
			\quad++Re-ranking & 40.5 & 49.1 & 0.427 \\
			\bottomrule
	\end{tabular}}
	\caption{EN$\rightarrow$ZH performance (charBLEU and chrF score) for different models.}
	\label{tab:enzh_bleu}
\end{table}

\begin{table*}[t]
	\centering
	\scalebox{1.0}{
		\begin{tabular}{lccccccc}
			\toprule
			\multirow{2}{*}{Systems} & \multicolumn{3}{c}{DE$\rightarrow$HSB} & & \multicolumn{3}{c}{HSB$\rightarrow$DE} \\
			\cmidrule{2-4}\cmidrule{6-8} & Dev & Test & Official & & Dev & Test & Official \\
			\midrule
			UnsupSMT \cite{artetxe2018emnlp} & 17.1 & 14.7 & - & & 13.8 & 12.6 & - 
			\\
			\midrule
			MASS baseline & 29.8 & 26.0 & - & & 31.4 & 27.3 & -\\
			\hdashline
			UNMT baseline & 31.1 & 27.2 & - & & 31.3 & 27.2 & - \\
			\quad+CLM finetune  & 29.2 & 25.6 & - & & 28.6 & 24.5 & - \\
			\quad+MLM finetune  & 32.4 & 28.3 & - & & 32.4 & 27.3 & - \\
			\quad+ESC finetune & 32.1 & 28.3 & - & & 32.2 & 27.8 & - \\
			\midrule
			EN-DE-HSB MUNMT baseline & 29.3 & 25.6 & - & & 30.0 & 26.2 & - \\
			\quad ++EN-DE NMT & 33.6 & 29.3 & - & & 33.6 & 29.6 & - \\
			\quad ++MLM finetune & 35.1 & 30.5 & 28.6 & & 34.9 & 30.7 & 28.6 \\
			\quad ++RAT + RABT + XBT & 47.8 & 41.8 & 40.3 & & 40.6 & 35.9 & 32.8 \\
			\bottomrule
			
	\end{tabular}}
	\caption{DE$\leftrightarrow$HSB unsupervised performance (sacreBLEU score) for different models.}
	\label{tab:dehsb_unsup_bleu}
\end{table*}

\paragraph{EN-ZH}
In EN-ZH, the pre-training of Longformer as a document encoder is unique. As described in \cite{Beltagy2020Longformer}, the Longformer needs a large number
of gradient updates to learn the local context first; before learning to utilize longer context. In the first phase of the staged training procedure, an initial RoBERTa \cite{liu2019roberta} model implemented in Fairseq \cite{ott2019fairseq} repository was trained on the sentence-level text available. In each subsequent phase, we trained the model on the paragraph text, doubled the window size and the sequence length, and halve the learning rate. For the paragraph text, the Wikidumps and NewsCommentary v15 have document intervals and can be used directly, while UN v1.0 has no document intervals but the sentence order is not interrupted. Therefore, we use the BERT Next Sentence Prediction (NSP) classification model provided by Google for document interval prediction to recover the documents.

\paragraph{DE-HSB}
In RUNMT on EN-DE-HSB, Europarl v10 EN-DE parallel corpus is used for EN-DE NMT and RAT/RABT/XBT training. Additionally, the BPE size increases to 50K for three languages. In CFST, the filtering threshold of BT-BTBLEU is set to $\gamma = 50.0$.

\begin{table*}[t]
	\centering
	\scalebox{1.0}{
		\begin{tabular}{lccccccc}
			\toprule
			\multirow{2}{*}{Systems} & \multicolumn{3}{c}{DE$\rightarrow$HSB} & & \multicolumn{3}{c}{HSB$\rightarrow$DE} \\
			\cmidrule{2-4}\cmidrule{6-8} & Dev & Test & Official & & Dev & Test & Official \\
			\midrule
			UNMT baseline & 31.1 & 27.2 & - & & 31.3 & 27.2 & - \\
			\quad++MLM finetune  & 32.4 & 28.3 & - & & 32.4 & 27.3 & - \\
			\quad++DE-HSB NMT & 59.9 & 53.0 & 52.5 & & 61.6 & 53.1 & 54.6 \\
			\quad++TLM finetune & 60.2 & 53.2 & - & & 61.4 & 52.7 & - \\
			\quad++CFST & 61.3 & 54.5 & 60.2 & & 62.2 & 53.9 & 55.6 \\
			\quad++D2GPo & 61.4 & 54.6 & 60.4 & & 62.9 & 54.5 & 56.6 \\
			\hdashline
			Ensemble+Re-ranking & 61.5 & 54.7 & 60.7 & & 63.3 & 56.1 & 58.5 \\
			\midrule
			EN-DE-HSB MUNMT baseline & 29.3 & 25.6 & - & & 30.0 & 26.2 & - \\
			\quad++EN-DE NMT + MLM finetune & 35.1 & 30.5 & 28.6 & & 34.9 & 30.7 & 28.6 \\
			\quad++DE-HSB NMT & 59.8 & 53.0 & - & & 62.0 & 53.7 & - \\
			\bottomrule
			
	\end{tabular}}
	\caption{DE$\leftrightarrow$HSB low-resource performance (sacreBLEU score) for different models.}
	\label{tab:dehsb_lowres_bleu}
\end{table*}

\section{Results and Analysis}

Results and ablations for PL$\rightarrow$EN\footnote{The team name for PL$\rightarrow$EN submission is ``NICT-rui" in the OCELoT site to distinguish between different sub-teams.} are shown in Table \ref{tab:plen_bleu}, EN$\rightarrow$ZH in Table \ref{tab:enzh_bleu}, unsupervised DE$\leftrightarrow$HSB in Table \ref{tab:dehsb_unsup_bleu} and low-resource DE$\leftrightarrow$HSB in Table \ref{tab:dehsb_lowres_bleu}. We report case-sensitive SacreBLEU scores using SacreBLEU \cite{post2018call} for EN-PL, DE-HSB, and BLEU based on characters for EN-ZH. In the results, ``+" means addition based on baseline, and ``++" means cumulative addition based on the previous one.

In PL$\rightarrow$EN, the introduction of ParaCrawl data improves the baseline performance on the dev dataset by about 4.2 BLEU. +D2GPo, XLM-enhanced NMT, Bidirectional NMT, and ensembling outperforms our strong baseline by 2 BLEU point. Finally, finetuning and reranking further gives another 0.5 BLEU.

For EN$\rightarrow$ZH, as with PL$\rightarrow$EN, we see similar improvements with +D2GPo, XLM-enhanced NMT, ensembling and reranking. We also observe that the addition of Document-enhanced NMT is much more substantial, improving single model performance by over 1.5 BLEU.

In the unsupervised track, we compared CLM, MLM, and Explicit Sentence Compression (ESC) pre-training approaches joint trained with BT in the second stage of UNMT, respectively, and found that MLM and ESC had similar effects and were stronger than CLM. Moreover, the pre-training baseline of MLM was stronger than that of MASS.  The combination of unsupervised training of DE-HSB and supervised training of EN-DE achieves the purpose of transfer learning, and the improvement is greater than 3 BLEU. Based on the conclusion of MLM and BT joint training on the UNMT Baseline, we also got a similar trend on the MUNMT system. In the final system, the enhancement of RAT+RABT+XBT brought a BLEU increase of 11.7 and 4.2, respectively.

In the low-resource track, the model in the unsupervised track is used as the pre-trained model, and DE-HSB NMT and BT are jointly trained. Due to the DE-HSB parallel corpus, we can not only use MLM for monolingual pre-training, but also use TLM for cross-lingual pre-training. The addition of CFST and D2GPo further improves the effect of the model, indicating that these contributions are orthogonal. In addition, comparing UNMT with MUNMT given a parallel corpus, we found that although MUNMT used more data, it did not bring about a large enough effect improvement, so we will leave it for future research.

\section{Conclusion}

This paper describes SJTU-NICT's submission to the WMT20 news translation task. For three typical scenarios, we adopt different strategies. In this work, we not only study the pre-trained language model to enhance MT, but also consider the impact of document information on translation. We considered both the way of converting document alignment into sentence alignment and the use of BERT's NSP to recover the structure of documents. In addition, transfer learning from supervision is taken into account in unsupervised translation, and various means are used to enhance low-resource translation. Our systems performed strongly among all the constrained submissions: we ranked 1st in PL$\rightarrow$EN, EN$\rightarrow$ZH, and DE$\rightarrow$HSB respectively, and stayed Top-3 for the HSB$\rightarrow$DE.

\bibliographystyle{acl_natbib}
\bibliography{wmt2020}

\end{document}